\documentclass[11pt]{article}

\usepackage[preprint]{acl}

\usepackage{times}
\usepackage{latexsym}
\usepackage[T1]{fontenc}
\usepackage[utf8]{inputenc}
\usepackage{microtype}
\usepackage{inconsolata}
\usepackage{soul}
\usepackage{url}
\usepackage{graphicx}
\usepackage{amsmath}
\usepackage{amsthm}
\usepackage{booktabs}
\usepackage{multirow}
\usepackage{algorithm}
\usepackage{algorithmic}

\title{CacheWeaver: Cache-Aware Evidence Ordering for Efficient Grounded RAG Inference}

\author{
Kaizhen Tan, Rong Gu, Mingyuan Li\\
Heinz College of Information Systems and Public Policy, Carnegie Mellon University, USA
}

\begin{document}

\maketitle

\begin{abstract}
Retrieval-Augmented Generation (RAG) improves factual grounding, but it
also lengthens prompts and raises prefill cost. Prefix caching in
serving engines such as vLLM reduces this cost only when requests share
the same token prefix. In grounded generation, however, adjacent queries
may retrieve overlapping evidence in different orders, so set overlap
does not become reusable prefix overlap. We present CacheWeaver, a
lightweight prompt-layer method for cache-aware evidence ordering. The
method keeps a prefix tree over recently served evidence sequences and
uses a greedy walk to place the most reusable prefix first, while leaving
the serving engine and retrieved evidence set unchanged. Across three
vLLM configurations the method lowers median time-to-first-token (TTFT)
by about 20--33\% relative to retrieval-order prefix caching, without
hurting answer quality in our QA tests. The greedy policy reaches 97.5\% of the
median TTFT gain from oracle ordering, which indicates that most reusable prefix
locality can be recovered by a simple scheduling layer between retrieval
and inference.
\end{abstract}

\section{Introduction}
\label{sec:intro}

Retrieval-Augmented Generation (RAG) grounds a large language model
(LLM) by adding retrieved passages to the prompt before
generation~\cite{lewis2020rag,gao2024ragsurvey}. The price is serving
latency: the retrieved context usually dominates the input length, so
prefill becomes the main bottleneck for interactive use.

Serving engines address part of this cost with prefix caching. Systems
such as vLLM and SGLang reuse previously computed key--value (KV) states
through Automatic Prefix Caching (APC) when a new request shares a token
prefix with an earlier one~\cite{vllm_apc,zheng2024sglang}. This works
well for repetitive prompts, but much less so in RAG, where retrieved
passages overlap in content yet appear in different orders.

That mismatch is the problem we study. Prefix caching operates at the
token level, while retrieval returns documents at the chunk level. Two
consecutive queries may retrieve nearly the same evidence, but once the
first document differs the shared token prefix breaks and most reusable
KV blocks are lost. Document overlap is common in RAG; prefix overlap is
rare.

Our idea is to change the document order rather than the serving engine.
If retrieved documents are reordered to follow a recently cached
sequence, the resulting prompt is more likely to match an existing
prefix and benefit from APC. We build CacheWeaver, a prompt-layer
scheduler that maintains a knowledge tree over recently served document
sequences and uses a greedy walk to place the longest likely-cached
prefix first. The remaining documents follow in retrieval order, so the
system falls back to the original ranking whenever reuse is unavailable.
The design rests on three working assumptions: consecutive queries share
some documents, reordering evidence does not change correctness in
extractive tasks, and the engine's cache state can be approximated by
recency of service. These hold in many deployments where queries cluster
around related topics. We explicitly evaluate the second assumption
because evidence order can affect grounded generation.

Recent work shows that context order affects cache reuse in long-context
and RAG serving~\cite{jin2025ragcache,cachecraft2025,contextpilot2025}.
Our goal is not to re-establish that ordering matters, but to ask how
much of the benefit a minimal, engine-agnostic design can recover. The
core policy stays entirely at the prompt-construction layer, requires no
change to vLLM, and still performs close to exhaustive oracle ordering.

We study three questions. Can a pure prompt-layer policy deliver
meaningful median time-to-first-token (TTFT) gains in APC-based grounded RAG
inference? Is a greedy trie traversal enough, or does the problem demand
heavier search? And do the median TTFT gains survive beyond a single synthetic
benchmark, across public question answering (QA) slices, overlap levels,
top-$k$ values, and concurrent execution? Our experiments answer yes on
locality-bearing workloads and also identify the boundary case: public
slices show speedups only when adjacent queries actually share retrieved
evidence.

Concretely, we make three contributions. First, we cast cache-aware
evidence ordering as a prompt-layer optimization that maximizes reusable
prefix depth while leaving the serving engine and the retrieved evidence
set unchanged. Second, we give a greedy trie scheduler for this
objective, with a complexity analysis and an argument for why it
approaches exhaustive ordering. Third, we evaluate it across three
measurement configurations and several workloads, including HotpotQA,
Natural Questions Open (NQ-Open), and TriviaQA public-data checks and a
small top-$k$ sweep.
The method gives large median TTFT gains on
locality-bearing traces and little median TTFT gain on no-locality public slices,
clarifying when cache-aware evidence ordering is useful, while matching
retrieval-order answer metrics in bounded QA checks.

\section{Background and Related Work}
\label{sec:background}

\subsection{RAG Serving and Prefix Caching}

A standard RAG pipeline encodes the query, retrieves documents,
constructs a prompt, and runs LLM inference. Prompt construction is
cheap; inference dominates latency. The serving problem is therefore not
how to concatenate passages but how to avoid repeated prefill work when
many requests share context.

vLLM manages KV-cache memory in fixed-size blocks through
PagedAttention~\cite{kwon2023pagedattention} and layers APC on
top~\cite{vllm_apc}. APC reuses cached blocks when a new request shares a
token prefix with an earlier one; each block is hashed together with its
prefix history, so reuse is inherently prefix-sensitive. Once an early
block differs, later blocks no longer line up with the cached chain even
when the two prompts contain the same documents, so a small difference
near the start of a prompt removes most potential reuse.

\begin{figure}[t]
\centering
\includegraphics[width=\linewidth]{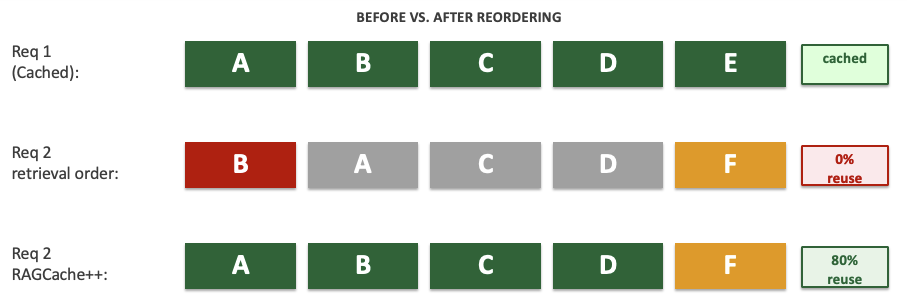}
\caption{Prefix alignment in APC-based RAG serving. Request~1 is cached.
Request~2 shares 80\% of its documents but, in retrieval order, the
prefix diverges at the first drawn document block, yielding no reusable
prefix. Reordering the shared documents to the front restores alignment.
For readability each drawn block represents one retrieval passage; in
vLLM a passage spans several fixed-size token blocks.}
\label{fig:prefix_blocks}
\end{figure}

Throughout the paper, a \emph{document} is the retrieval unit passed to
the prompt builder; it is also called a passage or chunk in some RAG
systems. In our benchmarks this unit is one retrieved passage: synthetic
passages contain about 200 tokens, while HotpotQA passages are truncated
to roughly 160 words to fit the context window. APC does not cache these
units directly. It caches fixed-size token blocks, so a matching document
sequence maps to the contiguous run of vLLM blocks produced by those
passage tokens. In our vLLM runs, the token block size is 16 tokens.
Figure~\ref{fig:prefix_blocks} therefore idealizes a document as one
drawn block for readability; the actual measurements use vLLM's
token-block APC unchanged.

This matters for RAG because retrieval overlap is set-based while APC
demands ordered prefix agreement. When two requests retrieve the same
documents in different positions, their token prefixes diverge
immediately and reuse collapses. Figure~\ref{fig:prefix_blocks}
illustrates the gap and shows why reordering is a natural intervention
point.

\subsection{Related Work}

Several recent systems study KV reuse in long-context and RAG inference
at different layers. \textbf{RAGCache}~\cite{jin2025ragcache} is closest
to ours: it also builds a tree over document sequences, but ties the
design to a richer cache-management stack inside the serving system. We
borrow the intuition of sequence-aware reuse and target a smaller
problem, evidence ordering at the prompt layer, with no engine change.
\textbf{ContextPilot}~\cite{contextpilot2025} reuses context for
long-context inference through a context index, ordering, and
de-duplication, and supports the same broad message: prefill can shrink
by reorganizing context rather than redesigning the model.
\textbf{Cache-Craft}~\cite{cachecraft2025} shows that plain prefix
caching weakens in production RAG because retrieved chunks and their
order vary across requests, which directly motivates our setting.
\textbf{TurboRAG}~\cite{turborag2024} and
\textbf{CacheBlend}~\cite{yao2025cacheblend} tackle the harder problem of
reusing KV even when reusable chunks are not one contiguous prefix; these
methods are more general but more involved than ours, which works with
existing APC. \textbf{CacheGen}~\cite{lmcache2024} compresses and streams
KV caches, which is orthogonal and could be combined with cache-aware
ordering. Table~\ref{tab:related_positioning} summarizes the design
point. We ask how much reuse is recoverable with prompt-layer ordering
alone on an unmodified prefix-caching engine, and we treat exhaustive
oracle ordering as the upper bound for that design point.

\begin{table*}[t]
\caption{Positioning relative to recent cache-reuse systems. The
extra-support column describes runtime or cache support beyond a normal
prefix-caching server.}
\label{tab:related_positioning}
\centering
\small
\resizebox{\textwidth}{!}{%
\begin{tabular}{@{}lllll@{}}
\toprule
\textbf{Method} & \textbf{Layer} & \textbf{Extra runtime/cache support} & \textbf{Reuse target} & \textbf{Main contrast} \\
\midrule
RAGCache~\cite{jin2025ragcache} & Serving cache manager & Cache-manager support & Document-sequence KV cache & Richer cache stack inside the serving system \\
CacheBlend~\cite{yao2025cacheblend} & Runtime KV fusion & KV-fusion support & Non-prefix chunk KV & Reuses cached chunks beyond shared prefixes \\
TurboRAG~\cite{turborag2024} & Offline KV precomputation & Offline KV-cache generation and retrieval & Chunk-level KV & Retrieves precomputed document-chunk KV caches \\
ContextPilot~\cite{contextpilot2025} & Context-reuse runtime & Runtime and context-index support & Reusable context segments & General long-context reuse, ordering, and de-duplication \\
Cache-Craft~\cite{cachecraft2025} & Chunk-cache management & Cache-manager support & Chunk-level KV caches & Manages reusable chunk caches for production RAG \\
CacheWeaver (ours) & Prompt construction & No engine/cache support beyond middleware & Ordered evidence prefix & Uses existing prefix caching by reordering evidence \\
\bottomrule
\end{tabular}
}
\end{table*}

\section{Method}
\label{sec:design}

\subsection{Overview}

CacheWeaver sits between retrieval and prompt construction. Given a
retrieved document set, it reorders the documents so the final prompt is
more likely to share a long prefix with recently served prompts. The
serving engine still performs all KV-cache allocation and reuse; our
layer only decides the document order.

\subsection{Knowledge Tree}

The knowledge tree is a trie whose root-to-leaf paths record document
orders that recently appeared in served requests. Each node stores a
document ID with lightweight metadata about recent use. The tree supports
two operations: after a request finishes, its ordered document sequence
is inserted; when a new request arrives, the trie is consulted to
estimate which order is most likely to align with a cached prefix. It
thus acts as a compact memory of recent prompt structure, not as a
replacement for vLLM's own KV cache.

\subsection{Greedy Ordering Algorithm}

\textbf{Problem.} Let $D=\{d_1,\dots,d_k\}$ be the retrieved documents in
retrieval rank, and let an ordering $O$ be a permutation of $D$ used to
build the prompt. The knowledge tree $T$ records recently served orders,
and a node is \emph{cached} when its KV blocks are predicted resident.
For an ordering $O$, define its \emph{reusable prefix depth} $\ell(O,T)$
as the length of the longest leading run $d_{O(1)},\dots,d_{O(\ell)}$ that
follows a cached root-to-node path in $T$. Because APC savings grow with
the shared prefix length, we seek
\begin{equation}
O^\star = \arg\max_{O\in\Pi(D)} \ell(O,T),
\end{equation}
where $\Pi(D)$ is the set of permutations of $D$, and documents past the
reused prefix keep their retrieval rank. Restricting $O$ to $\Pi(D)$ keeps
the top-$k$ evidence set intact, so reordering changes prefill cost
without changing what the model sees.

Algorithm~\ref{alg:ordering} gives the ordering policy. Starting from the
trie root, it repeatedly checks whether one remaining retrieved document
extends a cached path; if so, that document is placed next, and if no
cached continuation exists, the remaining documents are appended in
retrieval order.

\begin{algorithm}[t]
\caption{Cache-Aware Evidence Ordering}
\label{alg:ordering}
\begin{algorithmic}[1]
\REQUIRE Retrieved doc IDs $D = \{d_1, \ldots, d_k\}$ (in retrieval rank), knowledge tree $T$
\ENSURE Ordered sequence $O$
\STATE $O \leftarrow []$, $R \leftarrow D$, $\textit{node} \leftarrow T.\text{root}$
\WHILE{$R \neq \emptyset$}
    \STATE $\textit{found} \leftarrow \texttt{false}$
    \FOR{each $d \in R$}
        \STATE $c \leftarrow \textit{node}.\text{getChild}(d)$
        \IF{$c \neq \texttt{null}$ \AND $c.\text{isCached}()$}
            \STATE Append $d$ to $O$; remove $d$ from $R$
            \STATE $\textit{node} \leftarrow c$; $\textit{found} \leftarrow \texttt{true}$; \textbf{break}
        \ENDIF
    \ENDFOR
    \IF{$\textit{found} = \texttt{false}$}
        \STATE Append remaining $R$ in retrieval-rank order to $O$
        \STATE \textbf{break}
    \ENDIF
\ENDWHILE
\STATE \textbf{return} $O$
\end{algorithmic}
\end{algorithm}

The policy is intentionally simple: it runs no combinatorial search at
runtime and instead greedily extends the longest usable prefix reachable
from the current trie state. Each iteration probes at most $|R|$ trie
children, so the walk costs $O(k^2)$ child lookups for top-$k$ retrieval,
negligible for the small $k$ used in practice, and the tree costs
$O(Hk)$ space for $H$ retained request paths, bounded by the recency
window. An exhaustive oracle, by contrast, searches over the $k!$
candidate orders.

\begin{figure}[t]
\centering
\includegraphics[width=\linewidth]{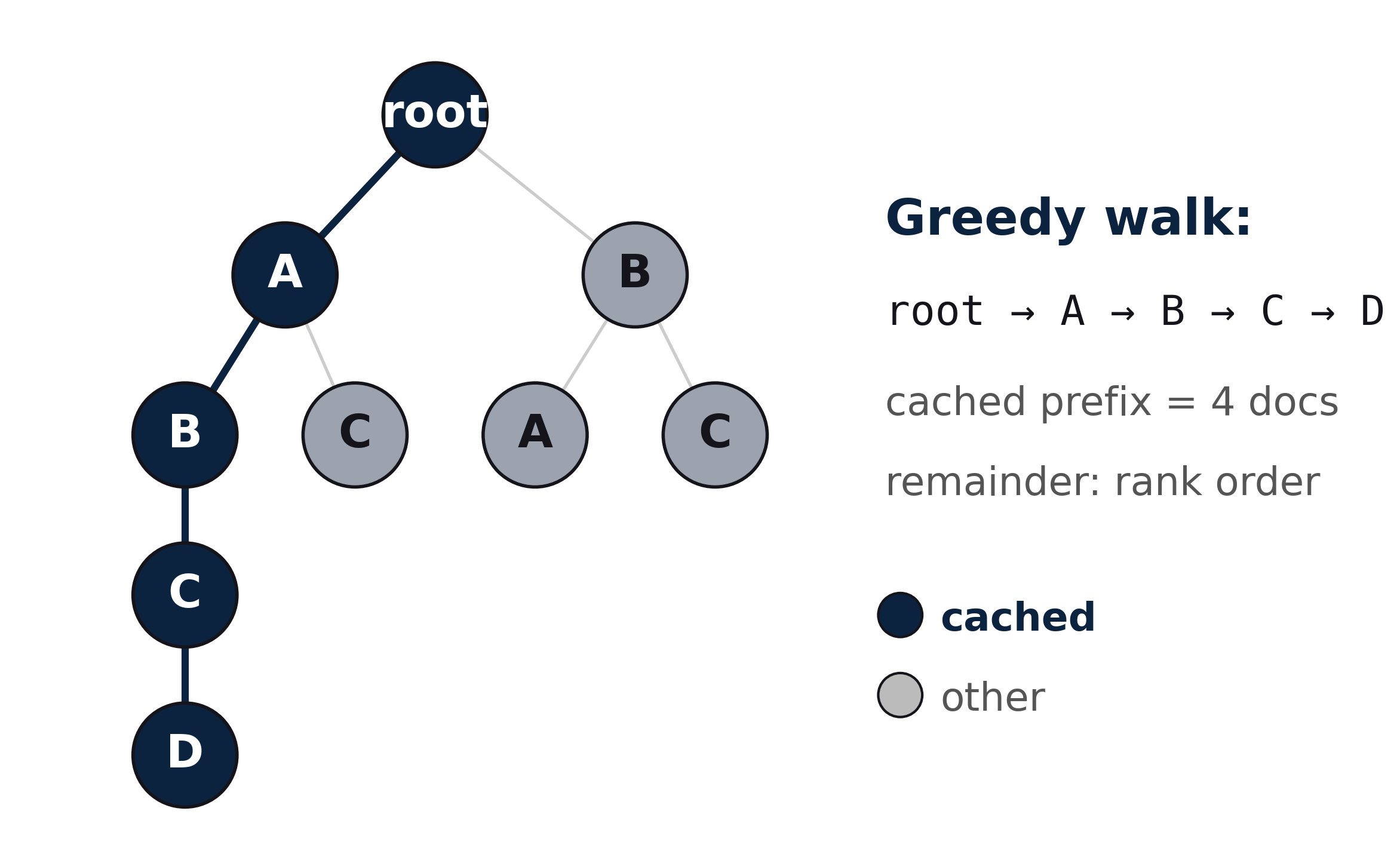}
\caption{Knowledge tree example. Dark nodes are cached; gray nodes are
evicted or on other branches. Given candidate set $\{A,B,C,F,D\}$, the
greedy walk follows root$\to$A$\to$B$\to$C$\to$D (the longest cached
path), yielding a 4-document prefix hit; the remaining document (F) is
appended in retrieval-rank order.}
\label{fig:trie}
\end{figure}

Greedy is optimal for the objective $\ell(O,T)$ whenever every visited
node has at most one cached child that still lies in $R$: with no branch
ambiguity, following the single cached continuation yields the deepest
reusable prefix. Sub-optimality can arise only when two cached children
remain retrievable at the same node, which is rare under the bursty
locality that motivates the method, since recent orders tend to recur.
This explains the result of Section~\ref{sec:analysis}, where greedy
ordering trails the oracle by only 0.46\,ms at p50
(Table~\ref{tab:baselines}). Figure~\ref{fig:trie} shows the intuition.

\subsection{Cache-State Approximation, Feedback, and Implementation}
\label{sec:feedback}

Unless stated otherwise, the headline experiments use the minimal
ordering path: the trie proposes an evidence order, and vLLM performs APC
normally. The knowledge tree only approximates the live APC state: it
records what was served recently, not what remains resident in GPU
memory, which vLLM controls through internal block eviction. The trie can
therefore predict reuse for a path whose blocks were already evicted. We
treat this as a systems issue rather than a modeling failure. In stable
sequential runs the approximation is often good enough on its own. For
tighter memory budgets or heavier concurrency, the implementation also
supports a TTFT-based feedback loop: when the trie predicts a long
reusable prefix but the observed TTFT stays close to a cold request, the
supposed cached path is probably gone, so we mark it stale and stop
relying on it. A secondary optimization can overlap retrieval with a
warmup of the shared system prompt, whose KV states can be prefetched
while the retriever runs; this mainly helps cold starts and recovery
after eviction and does not change the ordering policy.

We implement CacheWeaver as a Python middleware over vLLM and FastAPI. It
uses only the public \texttt{LLM.generate} API with APC enabled, so it
needs no private engine internals and stays close to a drop-in component
rather than a fork. The trie and prompt builder are decoupled from cache
bookkeeping, so the ordering logic runs and is unit-tested without a GPU.
The fallback is simple and safe: an empty trie or a missing cached
continuation returns retrieval order. Feedback is an implementation
option for tighter cache budgets rather than part of the headline
measurement path.

\section{Experimental Setup}
\label{sec:eval}

\paragraph{Hardware and models.} Table~\ref{tab:config} summarizes all
measurement configurations. The headline controlled TTFT experiments use
Configs~A and~B with Qwen2.5 models~\cite{qwen25tr}, while Config~C is
used for the public-data and top-$k$ sensitivity checks. Headline
experiments use top-$k=5$ retrieval because it is a common RAG setting
and keeps the oracle comparison exact ($5! = 120$ candidate orders per
request); we add a small $k \in \{3,5,7,10\}$ sensitivity sweep below.
These are separate measurement campaigns, so absolute TTFT values should
be compared within a table rather than across tables.

\begin{table}[t]
\caption{Measurement setup summary.}
\label{tab:config}
\centering
\small
\resizebox{\columnwidth}{!}{%
\begin{tabular}{@{}llll@{}}
\toprule
 & \textbf{Config A} & \textbf{Config B} & \textbf{Config C} \\
\midrule
GPU & RTX~4060\,Ti 8\,GB & RTX~4090 24\,GB & RTX~4090D 24\,GB \\
Model & Qwen2.5-1.5B & Qwen2.5-7B & Qwen2.5-7B \\
vLLM & 0.8.5.post1 & 0.17.1 & 0.8.5.post1 \\
\texttt{max\_model\_len} & 2048 & 4096 & 4096 \\
\texttt{gpu\_mem\_util} & 0.85 & 0.90 & 0.90 \\
\bottomrule
\end{tabular}%
}
\end{table}

\paragraph{Workload.} The main benchmark uses a controlled synthetic
corpus in which consecutive queries within a burst retrieve overlapping
documents. The 100-document and 500-document corpora are not subset
relations; they are independently generated from the same region-based
template, with roughly 200-token passages grouped into topical regions.
The vLLM versions differ because these configurations were run in
separate measurement campaigns: Config~A on a local 4060\,Ti setup,
Config~B for the headline controlled trace on an RTX~4090, and
Config~C for public-data and top-$k$ checks. All conclusions compare
strategies within the same configuration, not absolute latency across
vLLM versions. For two queries with retrieved sets $A$ and $B$ we measure
overlap by the Jaccard similarity $\lvert A \cap B\rvert / \lvert A \cup
B\rvert$. The setup is deliberately simple: it isolates the systems
question of whether reordering can turn document overlap into
APC-friendly prefix reuse.

\paragraph{Prompt format.} Every prompt is a fixed instruction, then the
ordered evidence blocks under \texttt{[Document $r$: id]}
headers, then the query. All evidence precedes the query, so requests
with different questions can still share a token prefix; the rank $r$ in
each header means that prefix requires the shared documents in the same
leading positions. Only the document order differs between strategies.
Appendix~\ref{sec:prompt} gives the verbatim templates.

\paragraph{Strategies.} We compare four settings: no cache, APC with
retrieval order, APC with lexicographic sorting, and APC with our
optimized ordering, each on a fresh model instance to avoid cache-state
contamination.

\paragraph{Metrics.} We focus on TTFT, since evidence ordering affects
prefill rather than decode. We also report Fast\%$^{\dagger}$, a
TTFT-based reuse proxy that our harness computes separately inside each
strategy run $s$:
\begin{equation}
\label{eq:fastpct}
\mathrm{Fast\%}^{\dagger}_{s}
= \frac{100}{N}\sum_{i=1}^{N}
  \mathbf{1}\!\left[t^{s}_{i} < 0.6\,t^{s}_{\mathrm{warm}}\right].
\end{equation}
Here $t^{s}_{i}$ is the post-warmup TTFT of request $i$ under strategy
$s$, and $t^{s}_{\mathrm{warm}}$ is the median TTFT of the five warmup
requests of that same run. Each strategy is normalized by its own early,
mostly cold requests, so the threshold differs per strategy and
Fast\%$^{\dagger}$ must be read within a strategy rather than compared
across strategies. TTFT stays the primary metric, and the cross-strategy
compute comparison uses the prefill-saved fraction of
Table~\ref{tab:cache_eff}, which divides every strategy by one shared
no-cache denominator.

\section{Main Results}
\label{sec:main}

\begin{figure}[t]
\centering
\includegraphics[width=0.92\linewidth]{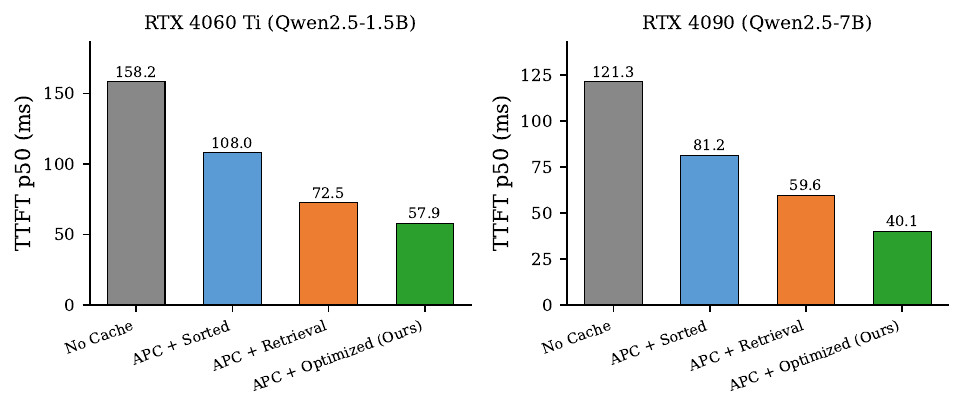}
\caption{Median TTFT across strategies. Optimized ordering attains the
lowest median TTFT on both headline configurations.}
\label{fig:ttft}
\end{figure}

Table~\ref{tab:main} shows the headline result, and
Figure~\ref{fig:ttft} visualizes the same median TTFT comparison:
optimized ordering consistently lowers median TTFT on both headline
configurations.

\begin{table}[t]
\caption{Headline TTFT results. TTFT is in ms. Best APC result is bold.
Fast\%$^{\dagger}$ (Eq.~\ref{eq:fastpct}) is a within-strategy reuse
proxy: a request counts as fast if its TTFT falls below 60\% of the
median warmup TTFT of the \emph{same} run. Each column entry is
normalized by its own strategy, so these values are not comparable across
rows.}
\label{tab:main}
\centering
\small
\resizebox{\columnwidth}{!}{%
\begin{tabular}{@{}llcccc@{}}
\toprule
\textbf{Setup} & \textbf{Strategy} & \textbf{p50} & \textbf{p95} & \textbf{Mean} & \textbf{Fast\%$^{\dagger}$} \\
\midrule
\multirow{4}{*}{\rotatebox{90}{\scriptsize Config A}}
 & No Cache      & 158.2 & 161.4 & 158.3 & 0.0 \\
 & APC+Retrieval & 72.5  & \textbf{159.9} & 69.9  & 17.9 \\
 & APC+Sorted    & 108.0 & 164.9 & 101.7 & 25.3 \\
 & \textbf{APC+Optimized} & \textbf{57.9} & 161.3 & \textbf{68.1} & 18.9 \\
\midrule
\multirow{4}{*}{\rotatebox{90}{\scriptsize Config B}}
 & No Cache      & 121.3 & 123.7 & 121.4 & 0.0 \\
 & APC+Retrieval & 59.6  & 117.4 & 55.9  & 12.3 \\
 & APC+Sorted    & 81.2  & 118.7 & 76.3  & 25.6 \\
 & \textbf{APC+Optimized} & \textbf{40.1} & \textbf{117.4} & \textbf{51.4} & 19.0 \\
\bottomrule
\end{tabular}%
}
\end{table}

Two points matter more than the raw numbers. First, the gain
concentrates at p50 rather than p95, as expected: tail cases are often
cold requests where no ordering policy can create reuse. Second,
optimized ordering beats lexicographic sorting by 46\% at p50 on
Config~A and by 51\% on Config~B, even though sorting also produces a
canonical, query-independent order. The Fast\%$^{\dagger}$ columns cannot
be used to explain this, because each is normalized by its own run
(Eq.~\ref{eq:fastpct}). The cross-strategy evidence that a few deep
prefix matches are worth more than many shallow ones is the prefill-saved
fraction in Table~\ref{tab:cache_eff}, where optimized ordering removes
66.3\% of median prefill against 50.1\% for retrieval order.
Tables~\ref{tab:main}, \ref{tab:baselines}, and \ref{tab:profiling}
come from separate runs for headline comparison, oracle comparison, and
profiling, respectively. The Config~B p50 values therefore differ slightly
across these tables because each run uses a fresh model process and a
measurement goal-specific trace; the authoritative comparison is the
within-table delta between strategies.

\section{Analysis}
\label{sec:analysis}

\paragraph{Greedy versus oracle ordering.} We compare the greedy walk
with alternative orderings, including a recency heuristic and an
exhaustive oracle (Table~\ref{tab:baselines}). The greedy method reaches
97.5\% of the oracle's TTFT reduction ($+29.6\%$ versus the oracle's
$+30.3\%$), so expensive search is unnecessary at this problem size.

\begin{table}[t]
\caption{Ordering baseline comparison on Config~B.}
\label{tab:baselines}
\centering
\small
\resizebox{\columnwidth}{!}{%
\begin{tabular}{@{}lccr@{}}
\toprule
\textbf{Strategy} & \textbf{p50 (ms)} & \textbf{Mean (ms)} & \textbf{vs.\ Retr.} \\
\midrule
No Cache & 121.55 & 121.73 & -- \\
APC+Sorted & 81.75 & 77.42 & $-35.5$\% \\
APC+Retrieval & 60.34 & 57.63 & baseline \\
APC+Recency & 59.52 & 56.62 & $+1.4$\% \\
APC+Optimized (Ours) & \textbf{42.50} & \textbf{53.77} & $+29.6$\% \\
APC+Oracle & 42.04 & 53.08 & $+30.3$\% \\
\bottomrule
\end{tabular}%
}
\end{table}

\begin{figure}[t]
\centering
\includegraphics[width=0.92\linewidth]{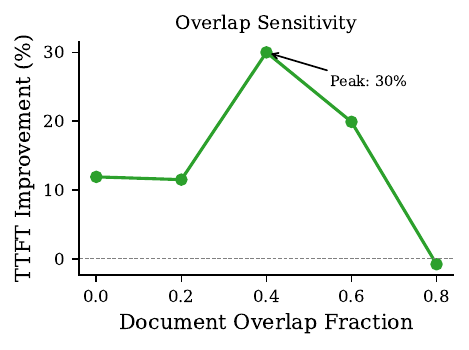}
\caption{TTFT improvement versus the trace generator's overlap setting.}
\label{fig:overlap}
\end{figure}

\paragraph{Top-$k$ sensitivity.} The retrieval depth controls both the
opportunity for prefix reuse and the cost of exploring document orders.
Table~\ref{tab:topk_sweep} gives a small sweep on the controlled trace.
Config~C uses 100 queries, with 95 measured after warmup. The gain
is not perfectly monotonic, but optimized ordering stays positive at all
tested $k$ values and becomes largest when the retrieved sets create
deeper reusable prefixes. Table~\ref{tab:topk_sweep} varies retrieval
depth, whereas Figure~\ref{fig:overlap} varies adjacent overlap, so the
two diagnostics isolate different sources of reusable prefix structure.

\begin{table}[t]
\caption{Top-$k$ sensitivity on Config~C.}
\label{tab:topk_sweep}
\centering
\small
\resizebox{\columnwidth}{!}{%
\begin{tabular}{@{}ccrrr@{}}
\toprule
\textbf{$k$} & \textbf{Mean Jac.} & \textbf{Retr. p50} & \textbf{Opt. p50} & \textbf{vs.\ Retr.} \\
\midrule
3  & 0.296 & 58.7 & \textbf{58.0} & $+1.2$\% \\
5  & 0.522 & 62.6 & \textbf{43.7} & $+30.2$\% \\
7  & 0.615 & 71.8 & \textbf{67.7} & $+5.7$\% \\
10 & 0.909 & 85.7 & \textbf{32.2} & $+62.4$\% \\
\bottomrule
\end{tabular}%
}
\end{table}

\paragraph{Overlap sensitivity.} Varying the document overlap in the
trace gives an intuitive pattern (Figure~\ref{fig:overlap}): ordering
helps most at moderate overlap; at very high overlap retrieval order is
already close to optimal; at near-zero overlap little reusable structure
exists. CacheWeaver is therefore not a universal speedup but is most
effective when requests share enough evidence to reuse, yet vary enough
that retrieval order is no longer cache-friendly.

\paragraph{Public-data locality diagnostics.} We add three public validation
workloads as locality and latency diagnostics
(Table~\ref{tab:public_workloads}). For HotpotQA
fullwiki~\cite{yang2018hotpotqa}, each query uses up to five
dataset-provided context passages, with supporting-fact titles first and
distractors kept in dataset order. This slice has almost no adjacent
reuse (995 unique passages among 996 retrieved passages), so it is a
no-locality sanity check: optimized ordering has no median TTFT gain. For
NQ-Open~\cite{kwiatkowski2019nq}, the public split contains questions
and answers but no passage contexts, so we build a term
frequency--inverse document frequency (TF-IDF) latency workload over
public question-answer records and do not report passage-grounded answer
quality. For TriviaQA~\cite{joshi2017triviaqa}, we use the
\texttt{rc.wikipedia} split and retrieve up to five Wikipedia evidence
chunks per question, placing chunks that contain answer aliases first and
then keeping the remaining evidence chunks in dataset order. HotpotQA and
NQ-Open use the first 200 validation examples in dataset order; TriviaQA
uses the first 200 validation examples after filtering examples with at
least one Wikipedia evidence chunk. In all three runs, the first five
requests are warmup, leaving 195 measured queries.

These fixed slices separate two requirements that a stronger public
experiment would combine. NQ-Open gives public traffic with modest
record-level locality and a 14.0\% median TTFT gain, but it lacks passage
contexts for our grounded QA check. HotpotQA and TriviaQA do provide
passage contexts, but their validation splits are independent samples
rather than chronological sessions, so retrieved passages almost never
recur. The knowledge tree then holds no cached continuation,
Algorithm~\ref{alg:ordering} falls back to retrieval rank on its first
iteration, and the optimized prompt is identical to the retrieval-order
prompt: on those two slices the method is a no-op by construction, the
residual 0.2--1.4\,ms differences in Table~\ref{tab:public_workloads} are
run-to-run noise, and answer quality is unchanged because the prompts are
the same. We therefore do not claim a public trace that simultaneously
shows locality, a median TTFT gain, and matched answer metrics; obtaining
one needs a sessionized public trace, which we leave to future work.

\begin{table}[t]
\caption{Fixed public-data locality and latency diagnostics on Config~C.
These are fixed validation slices, not natural chronological traffic.
TTFT is median ms.}
\label{tab:public_workloads}
\centering
\small
\resizebox{\columnwidth}{!}{%
\begin{tabular}{@{}llll@{}}
\toprule
\textbf{Dataset} & \textbf{Locality source} & \textbf{Mean Jac.} & \textbf{TTFT Ret$\rightarrow$Opt} \\
\midrule
HotpotQA & Dataset order & 0.000 & 97.9$\rightarrow$98.1 \\
NQ-Open & TF-IDF QA-record reuse & 0.068 & 34.3$\rightarrow$\textbf{29.5} \\
TriviaQA & Dataset order & 0.000 & 144.9$\rightarrow$146.3 \\
\bottomrule
\end{tabular}%
}
\end{table}

\paragraph{Answer-Quality Invariance Checks.} We report paired QA
metrics for retrieval order versus optimized order with the same evidence
set (Table~\ref{tab:grounding}). The controlled extractive and two-hop
checks report exact match (EM), contains-EM, and token F1. Contains-EM is
a relaxed exact-match metric that counts a prediction as correct when the
normalized gold answer appears anywhere in the generated response. We do
not report citation-support rate because these prompts ask for short
answers rather than evidence IDs. This is a narrow claim: position can
affect long-context QA quality and ``lost-in-the-middle''
behavior~\cite{liu2024lostmiddle}. Our evidence only shows that
retrieval order and optimized order match on these bounded checks; it
does not prove that evidence order never matters.

\begin{table}[t]
\caption{Bounded answer-quality checks with the same evidence set under
retrieval order and optimized order. Contains-EM counts a response as
correct when the normalized gold answer appears anywhere in the output.}
\label{tab:grounding}
\centering
\small
\resizebox{\columnwidth}{!}{%
\begin{tabular}{@{}llccc@{}}
\toprule
\textbf{Check} & \textbf{Order} & \textbf{EM} & \textbf{Contains EM} & \textbf{F1} \\
\midrule
Extractive synthetic & Retrieval & 1.000 & 1.000 & 1.000 \\
Extractive synthetic & Optimized & 1.000 & 1.000 & 1.000 \\
Two-hop natural & Retrieval & 0.675 & 0.775 & 0.862 \\
Two-hop natural & Optimized & 0.675 & 0.775 & 0.862 \\
Two-hop reversed & Retrieval & 0.625 & 0.750 & 0.826 \\
Two-hop reversed & Optimized & 0.625 & 0.750 & 0.826 \\
\bottomrule
\end{tabular}
}
\end{table}

\paragraph{Systems cost.} The middleware adds little measured host-side overhead.
As shown in Table~\ref{tab:profiling}, profiling on Config~B with 195
post-warmup queries shows that greedy ordering plus prompt construction
adds about 26\,$\mu$s per request (roughly 67\,$\mu$s total) while
reducing inference p50 by 29\% relative to retrieval order.
Throughput stays within 0.4\% of baseline APC in sequential and batched
execution (Table~\ref{tab:throughput}), so the median TTFT gain does not
reduce serving capacity. Because ordering uses the existing APC pool
rather than adding a separate GPU-resident cache, the knowledge tree
lives in host memory and stays small.

\begin{table}[t]
\caption{Ordering overhead on Config~B. This profiling run uses 195 post-warmup
queries. Optimized order plus build totals 67\,$\mu$s
($\sim$26\,$\mu$s over retrieval order) while cutting inference p50 by
29\% relative to retrieval order; OH is the host-side ordering overhead
as a fraction of total request time.}
\label{tab:profiling}
\centering
\small
\resizebox{\columnwidth}{!}{%
\begin{tabular}{@{}lccccc@{}}
\toprule
\textbf{Strategy} & \textbf{Order} & \textbf{Build} & \textbf{Gen.\ (mean)} & \textbf{Gen.\ (p50)} & \textbf{OH} \\
\midrule
APC+Retr. & 4.0\,$\mu$s & 36.3\,$\mu$s & 55.5\,ms & 58.6\,ms & 0.07\% \\
APC+Opt.  & 22.2\,$\mu$s & 44.4\,$\mu$s & 52.5\,ms & 41.4\,ms & 0.13\% \\
\bottomrule
\end{tabular}%
}
\end{table}

\begin{table}[t]
\caption{Throughput on Config~B. This run uses 200 queries with 50 output
tokens each; tok/s counts prompt plus output tokens. Optimized ordering matches
baseline APC within 0.4\%.}
\label{tab:throughput}
\centering
\small
\begin{tabular}{@{}lcccc@{}}
\toprule
 & \multicolumn{2}{c}{\textbf{Sequential}} & \multicolumn{2}{c}{\textbf{Batched}} \\
\cmidrule(lr){2-3}\cmidrule(lr){4-5}
\textbf{Strategy} & \textbf{req/s} & \textbf{tok/s} & \textbf{req/s} & \textbf{tok/s} \\
\midrule
No Cache      & 1.10 & 1{,}270  & 7.88  & 9{,}064  \\
APC+Retrieval & 1.19 & 1{,}370  & 82.05 & 94{,}382 \\
APC+Optimized & 1.19 & 1{,}371  & 81.75 & 94{,}029 \\
\bottomrule
\end{tabular}
\end{table}

\paragraph{Robustness diagnostics.} A second set of diagnostics tests
less controlled serving conditions. On
a diverse Wikipedia-style synthetic corpus with TF-IDF retrieval and no
manually imposed burst overlap, Table~\ref{tab:real_corpus} shows that
optimized ordering still lowers median TTFT by 26.8\% over retrieval
order. Although the mean adjacent Jaccard is low, repeated
topic-specific passages still create occasional reusable prefixes, which
the median TTFT captures. The median TTFT gain also persists under
concurrent batching, shrinking but staying positive across all tested
batch sizes (Table~\ref{tab:concurrent}), and narrows, though it stays
positive, as retrieval latency takes a larger share of end-to-end time
(Table~\ref{tab:e2e}).

\begin{table}[t]
\caption{TF-IDF corpus robustness on Config~B. This run uses 200 queries
and 1{,}000 passages over 20 topics, without manually imposed burst
overlap. Mean adjacent Jaccard is 0.06, and TTFT is in ms.}
\label{tab:real_corpus}
\centering
\small
\begin{tabular}{@{}lcccc@{}}
\toprule
\textbf{Strategy} & \textbf{p50} & \textbf{p95} & \textbf{Mean} & \textbf{vs.\ Retr.} \\
\midrule
No Cache      & 45.8 & 54.1 & 47.2 & -- \\
APC+Retrieval & 33.9 & 46.8 & 34.2 & baseline \\
APC+Optimized & \textbf{24.8} & \textbf{46.5} & \textbf{27.2} & +26.8\% \\
\bottomrule
\end{tabular}
\end{table}

\begin{table}[t]
\caption{Concurrent-load robustness on Config~B. TTFT is average per-request
latency in ms over 200 queries. The ordering benefit persists across all
concurrency levels.}
\label{tab:concurrent}
\centering
\small
\resizebox{\columnwidth}{!}{%
\begin{tabular}{@{}ccccr@{}}
\toprule
\textbf{Batch} & \textbf{No Cache} & \textbf{APC+Retr.} & \textbf{APC+Opt.} & \textbf{Improv.} \\
\midrule
1 & 114.7 & 54.7 & 50.7 & 7.3\% \\
2 & 110.5 & 48.0 & 44.3 & 7.7\% \\
4 & 103.4 & 41.6 & 37.5 & 9.9\% \\
8 & 100.9 & 38.3 & 35.7 & 6.8\% \\
\bottomrule
\end{tabular}%
}
\end{table}

\begin{table}[t]
\caption{End-to-end TTFT with retrieval latency on Config~B. This run uses
200 queries and reports TTFT in ms. The median TTFT benefit shrinks as retrieval takes a
larger share of total time but stays positive throughout.}
\label{tab:e2e}
\centering
\small
\begin{tabular}{@{}ccccr@{}}
\toprule
$T_{\text{ret}}$ & \textbf{Batch} & \textbf{APC+Retr.} & \textbf{APC+Opt.} & \textbf{Improv.} \\
\midrule
\multirow{3}{*}{10\,ms} & 1 & 69.6 & 51.8 & 25.6\% \\
 & 4 & 51.8 & 46.9 & 9.5\% \\
 & 8 & 48.9 & 45.9 & 6.1\% \\
\midrule
\multirow{3}{*}{30\,ms} & 1 & 89.7 & 72.4 & 19.3\% \\
 & 4 & 71.8 & 67.8 & 5.6\% \\
 & 8 & 68.8 & 66.8 & 2.9\% \\
\midrule
\multirow{3}{*}{50\,ms} & 1 & 109.3 & 92.2 & 15.6\% \\
 & 4 & 91.8 & 87.4 & 4.8\% \\
 & 8 & 88.7 & 86.3 & 2.7\% \\
\bottomrule
\end{tabular}
\end{table}

\paragraph{Compute efficiency.} TTFT gains align with reduced prefill
work. On Config~B, optimized ordering avoids 66.3\% of median per-request
prefill versus 50.1\% for retrieval-order APC, and benefits 185/195
requests versus 176/195. We use this prefill-saved fraction as a compute
proxy, not an energy or carbon measurement (Table~\ref{tab:cache_eff}).

\begin{table}[t]
\caption{Cache-efficiency proxy on Config~B (195 queries). Optimized
ordering saves deeper prefill on more requests than retrieval order.}
\label{tab:cache_eff}
\centering
\small
\resizebox{\columnwidth}{!}{%
\begin{tabular}{@{}lcccc@{}}
\toprule
\textbf{Strategy} & \textbf{Saved (p50)} & \textbf{Saved (mean)} & \textbf{Saved (p95)} & \textbf{Benefit} \\
\midrule
APC+Retr. & 50.1\% & 52.8\% & 75.6\% & 176/195 \\
APC+Opt.  & 66.3\% & 57.6\% & 77.7\% & 185/195 \\
\bottomrule
\end{tabular}%
}
\end{table}

\section{Conclusion}
\label{sec:conclusion}

CacheWeaver reorders retrieved evidence so prompts are more likely to
match recently cached prefixes, improving KV reuse without changing the
serving engine. In APC-based RAG, cache efficiency depends on the
evidence order as well as on the model and the backend, which makes
evidence ordering a lightweight scheduling problem between retrieval and
serving. A simple greedy
trie walk reduces median TTFT under locality-bearing traces, recovers
much of the lost locality, stays close to oracle ordering, adds
negligible host-side overhead, and matches retrieval-order answer metrics
in our bounded QA checks.

\paragraph{Limitations.} Headline traces are synthetic and bursty, and
public-data coverage is limited to fixed 200-example slices rather than a
sessionized public trace that jointly reports latency and bounded QA
metrics, so absolute median TTFT gains need not transfer unchanged to
production traffic. The benefit depends on temporal locality and moderate
document overlap: at very high overlap retrieval order often suffices,
and at very low overlap little reusable structure remains. Although our
bounded QA checks match retrieval-order quality, ordering could affect
tasks driven by positional attention patterns, reranker order, or
structured reasoning chains. We fix top-$k$ at 5 in the headline vLLM
experiments to keep oracle search tractable, so
Table~\ref{tab:topk_sweep} is an initial $k \in \{3,5,7,10\}$ sweep
rather than a production retrieval-depth study. The knowledge tree only
approximates true cache state, and the optional TTFT-based feedback loop
remains an indirect signal of residency. CacheWeaver fits
deployment-specific serving, such as customer-service systems and
enterprise knowledge bases, rather than general assistants handling
uncorrelated queries, and our evaluation is small in scale: a few models
and retrieval settings, and no large multi-tenant or production traces.

\paragraph{Future work.} The next step is to evaluate CacheWeaver on real
session traces, where
temporal locality comes from user behavior rather than controlled construction.
When serving engines expose direct cache-state signals, the middleware can use
them to align the knowledge tree with actual residency rather than relying only
on prompt history.

\paragraph{Societal impact.} This work targets serving efficiency rather
than application-level decision making, and it does not change the
retrieved evidence set, model, or safety policy. Lower RAG serving cost
can, however, make automated customer-service and decision-support
systems cheaper to deploy at scale; such systems should retain
domain-appropriate human oversight, logging, access control, and
grounding checks. The bounded answer-quality checks here are not a
general safety guarantee for downstream applications.

\section*{Acknowledgments}

We thank the anonymous reviewers for their comments. This work received
no external funding, and the authors declare no competing interests.

\bibliography{references}

\clearpage
\appendix

\section{Prompt Templates}
\label{sec:prompt}

The prompt builder concatenates a fixed instruction, one block per
retrieved document in the order chosen by the strategy, and the query.
The templates below are the exact strings used in all vLLM measurements;
line breaks inside the instruction are shown only to fit the column, and
the instruction itself is a single unbroken line in the code. Placeholders
in braces are substituted verbatim.

Latency runs (\texttt{max\_tokens=1}) use:

\begin{verbatim}
You are a helpful assistant. Answer
the question based only on the
provided documents. If the documents
don't contain enough information,
say so.

[Document 1: {doc_id}]
{document text}

[Document 2: {doc_id}]
{document text}

...

Question: {query}
Answer:
\end{verbatim}

\noindent
Answer-quality runs replace the instruction and the final cue so the
model emits a short span, keeping the evidence blocks unchanged:

\begin{verbatim}
Answer the question using only the
provided documents. Return the
shortest answer span, without
explanation.

[Document 1: {doc_id}]
{document text}

...

Question: {question}
Short answer:
\end{verbatim}

\noindent
Three properties of this layout decide whether cross-query prefix reuse
is possible. First, the instruction is a shared constant, so every
request begins with the same tokens. Second, all evidence precedes the
query, so a token prefix can be shared by requests with different
questions; a layout that placed the query first would make prefix reuse
impossible regardless of evidence ordering. Third, the block header
contains the retrieval rank $r$, so an evidence set reordered into
different positions produces different tokens. A cache hit therefore
requires the shared documents to appear in the same leading positions,
which is precisely the objective $\ell(O,T)$ that
Algorithm~\ref{alg:ordering} maximizes. Document identifiers are stable
strings: synthetic corpora use generated IDs, and public slices use an
8-hex-digit SHA-1 prefix of the passage title or chunk key, so the same
passage yields the same header across requests.

\end{document}